# Hybrid LSTM-Transformer Models for Profiling Highway-Railway Grade Crossings

Kaustav Chatterjee [1], Joshua Q. Li [2], Fatemeh Ansari [3], Masud Rana Munna [3], Kundan Parajulee [3], Jared Schwennesen [4]

[1] PhD, School of Civil Engineering, Oklahoma State University, Stillwater OK 74078; email: kaustav.chatterjee10@okstate.edu; orcid id: 0000-0002-9493-8694
[2] Professor/Williams Professorship, School of Civil Engineering, Oklahoma State University, Stillwater OK 74078; email: qiang.li@okstate.edu
[3] Graduate Research Associate, School of Civil Engineering, Oklahoma State University, Stillwater OK 74078
[4] Multimodal Division Manager, Multimodal Division, Oklahoma Department of Transportation, Oklahoma, OK 73105

## ABSTRACT

Hump crossings, or high-profile Highway Railway Grade Crossings (HRGCs), pose safety risks to highway vehicles due to potential hang-ups. These crossings typically result from post-construction railway track maintenance activities or non-compliance with design guidelines for HRGC vertical alignments. Conventional methods for measuring HRGC profiles are costly, time-consuming, traffic-disruptive, and present safety challenges. To address these issues, this research employed advanced, cost-effective techniques and innovative modeling approaches for HRGC profile measurement. A novel hybrid deep learning framework combining Long Short-Term Memory (LSTM) and Transformer architectures was developed by utilizing instrumentation and ground truth data. Instrumentation data were gathered using a highway testing vehicle equipped with Inertial Measurement Unit (IMU) and Global Positioning System (GPS) sensors, while ground truth data were obtained via an industrial-standard walking profiler. Field data was collected at the Red Rock Railroad Corridor in Oklahoma. Three advanced deep learning models—Transformer-LSTM sequential (model 1), LSTM-Transformer sequential (model 2), and LSTM-Transformer parallel (model 3)—were evaluated to identify the most efficient architecture. Models 2 and 3 outperformed the others and were deployed to generate 2D/3D HRGC profiles. The deep learning models demonstrated significant potential to enhance highway and railroad safety by enabling rapid and accurate assessment of HRGC hang-up susceptibility.

**INTRODUCTION**

High-profile Highway Railway Grade Crossing (HRGC) or hump crossing is an intersection where the profile of the road surface across the railway track can create the risk of highway vehicles getting stuck. Such intersections pose challenges for vehicles with long wheelbases, low ground clearance, or extended overhangs, as these vehicles may become "hung up" or high-centered on HRGCs. School buses, for instance, are susceptible to rear overhang-related hang-ups. Furthermore, trucks and tankers transporting hazardous materials are of critical concern at hump crossings due to the potentially catastrophic consequences. Various types of farm vehicles are also at risk of getting stuck. Additionally, high-profile crossings with steep-grade highway approaches could present difficulties for heavy vehicles, which may be required to stop at the crossings due to low acceleration capabilities and high gross weights.

Due to the profile of a hump crossing, the primary safety concern is the potential for collisions between highway vehicles and trains, which can lead to loss of life, injuries, property damage, and infrastructure damage. In 2015, a limousine was stuck on a hump crossing in New Paris, Indiana, and was struck by the Norfolk Southern Train (Liu et al. 2017). On October 15, 2021, a severe collision occurred between an Amtrak passenger train and a vehicle hauler truck high-centered at a hump railroad crossing near Thackerville, Oklahoma. Five train passengers were hospitalized with minor injuries (abcNEWS. 2021).

The risk of low-ground-clearance vehicles getting hung up on hump crossings can be mitigated by adhering to proper design guidelines for the vertical alignments of such HRGCs. These guidelines have been established and adopted by the American Association of State Highway and Transportation Officials (AASHTO, 2018), the American Railway Engineering and Maintenance-of-Way Association (AREMA, 2018), and the U.S. Department of Transportation (Ogden and Cooper, 2019). The guideline of recommended vertical HRGC alignment is illustrated in **Fig. 1**. However, even when new crossings are initially constructed following these design guidelines, maintenance activities on the railway track can alter the vertical alignment, potentially leading to the formation of high-profile HRGCs.

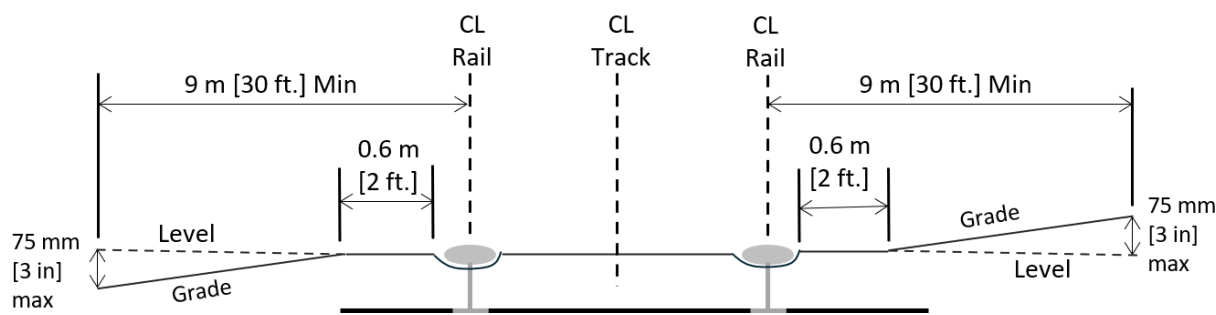

**Fig. 1.** Design Guidelines for vertical alignment of HRGCs (Ogden and Cooper. 2019, AASTHO. 2018, AREMA. 2018)

The issue of low-ground-clearance vehicles becoming hung up on HRGCs has been recognized by several researchers (Eck and Kang, 1991; 1992), which has led to the development of various techniques for determining HRGC profiles. Profile acquisition methods include physical models (Mutabazi and Russell, 2003) and Light Detection and Ranging (LiDAR) techniques (Khattak et al., 2015; Lee and Khattak, 2016; Liu et al., 2017). While these methods have proven beneficial, they also have notable drawbacks. The physical model proposed by Mutabazi and Russell (2003) is heavy, difficult to transport, and requires traffic control during data collection. LiDAR-based methods (Khattak et al., 2015; Lee and Khattak, 2016; Liu et al., 2017) are expensive, which demand specialized operators, and involve complex data analysis techniques. Furthermore, in many central U.S. regions, high-speed winds can restrict the use of LiDAR at certain times.

With the rapid advancement of sensing technologies and deep learning architectures, innovative approaches have become feasible for efficiently and accurately determining three-dimensional HRGC profiles. This research developed a hybrid deep learning model to estimate HRGC profiles by integrating Long Short-Term Memory (LSTM) and Transformer architectures. Training data were collected from various HRGCs in Oklahoma using advanced instruments, including the Pave3D8k laser imaging system for instrumentation data and a walking profiler for ground truth profile data.

A sequence-to-sequence LSTM-Transformer model was developed to leverage the Inertial Measurement Unit (IMU) and Global Positioning System (GPS) sensor data as inputs to generate precise HRGC profile outputs.

Previous studies have demonstrated the utility of IMU and GPS measurements for analyzing road geometry and dynamic parameters, such as roll and pitch (Bae et al., 2001; Jauch et al., 2017; Guang et al., 2021; Narkhede et al., 2021). For instance, Bae et al. (2001) used GPS sensors to determine road grade, while Jauch et al. (2017) employed IMU data - comprising accelerometer, gyroscope, and magnetometer readings - along with an orientation filter to calculate road grade and vehicle pitch. Guang et al. (2021) utilized an LSTM model to fuse IMU and GPS data for vehicle position estimation and demonstrated the model's effectiveness and applications in integrating these data sources.

**HRGC PROFILE FIELD DATA COLLECTION**

HRGC profile data acquisition was performed at the RedRock Corridor in Oklahoma, with a total of 225 HRGCs. **Fig. 2** presents some descriptive statistics about the crossings situated on the RedRock Corridor. Around 57% of the crossings are located on low-volume roads with less than 500 vehicles daily. Moreover, 76.0% of the HRGCs are paved, while 23.5% are unpaved and the median percentage of trucks traversing across crossings is around 5%. The vertical HRGC profiles were acquired using the Pave3D8k Laser Imaging data vehicle to acquire instrumentation data at traveling speed, and an industry-standard SurPro walking profiler for ground truth measurements (**Fig. 3**).

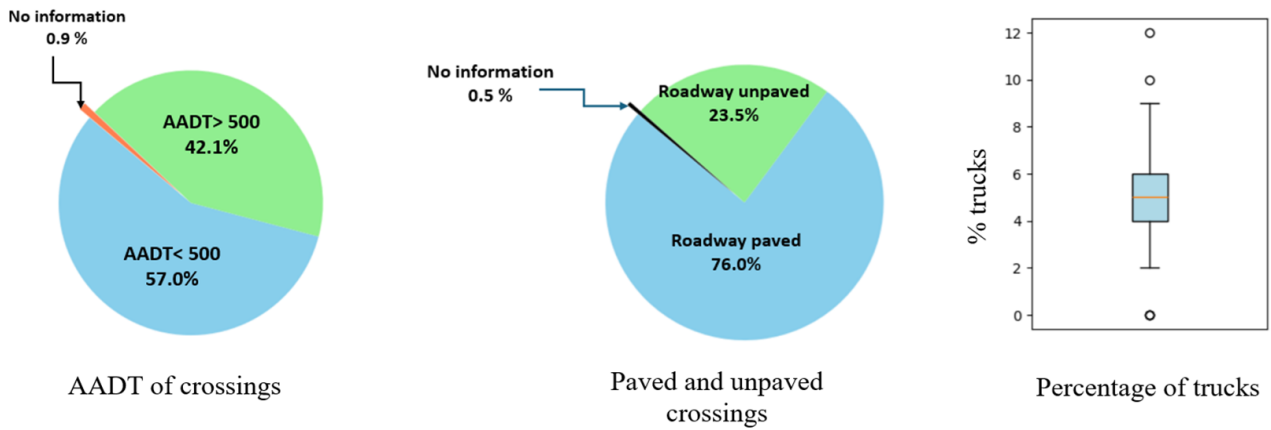

**Fig. 2**. Descriptives of BNSF RedRock Corridor

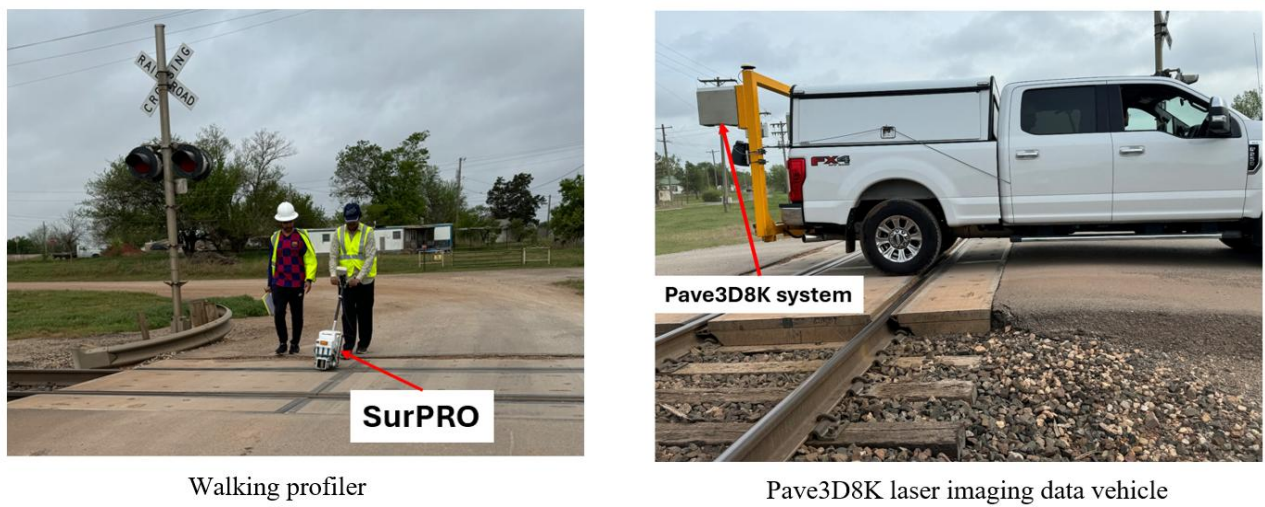

**Fig. 3**. Field data collection instruments

*Ground-Ttruth Profiles: the SurPro Walking Profiler*

The SurPro walking profiler is a well-recognized "golden" standard profile measuring instrument that has been widely deployed to evaluate the profile of different kinds of surfaces, including the highway, and airfield pavement. This device performs profiling using an inclinometer and can acquire the profile data with high precision and reproducibility. This device is equipped with co-linear wheels (spacing of wheel: 250 mm), which enable accurate profile acquisition from the HRGC at a resolution of 1 cm. The output from the device can be extracted in different formats including PPF, PRO, and ERD.

Some limitations of SurPro include the slow speed of data collection, the requirement of traffic regulation, and risks to personnel collecting data on the HRGC. These limitations can be overcome by determining the pavement profile using Pave3D8K Laser Imaging data vehicle data collected at traffic speed.

*Instrumented Measurements: Pave3D8K Laser Imaging Data Vehicle*

The Pave3D8K Laser Imaging data vehicle can acquire pavement surface condition data at high resolution (sub-mm) (Chatterjee et al. 2024). This technology can operate at traffic speed; therefore, no traffic control is required. The different sensors equipped in this system include the IMU, GPS sensor, 3D cameras, laser imaging, and distance measuring instruments. The data from the IMU and GPS sensor were utilized to estimate the profile of HRGCs.

An IMU is a solid-state or electromechanical device equipped with various sensors (accelerometer, gyroscope, and magnetometer) for measuring the motion of an object. Accelerometers measure linear acceleration along the three different axes, and gyroscopes measure the angular rate (change in angular velocity) along the three different axes (Bosman. 2015). The IMU equipped in this system can measure acceleration and other data at a frequency rate of around 100 Hz.

The IMU-GPS sensor was employed for measuring the acceleration of the vehicle in three dimensions, roll, pitch, speed of vehicle, and latitude, longitude, and altitude of the vehicle. The altitude data from GPS sensors can be used to find out the profile of HRGC. However, it was observed there was a difference between the measured profile of HRGC using the GPS sensor and using the walking profiler. The difference in measured profile was significant for estimating the hang-up potential of hump crossings. The primary reason for the difference in measurement (profile) between the walking profiler and the GPS sensor was due to the dynamics of the vehicle near HRGC. This problem was solved by employing sequence-to-sequence deep learning models.

**DATA PREPARATION**

*Data Description*

This research deals with sequence-to-sequence modeling, where the input sequence is composed of seven different parameters, and the output sequence is comprised of one parameter. The different parameters of the input sequence were acceleration of the vehicle along the X, Y, and Z direction, roll, pitch, speed of the vehicle, and altitude of the vehicle (GPS). The output sequence was the profile from the walking profiler (ground-truth data). The ground-truth data was used for developing the deep learning model.

A noteworthy point about this sequential data is the dimensionality of the data. On average, the number of data collected from one HRGC was around 2500 points, so each HRGC sequence has 2500 rows and seven columns from the IMU-GPS sensor and 2500 rows and one column from the walking profiler. Another vital fact about the dataset is that data points within a single HRGC sequence are dependent on each other. However, each HRGC sequence considered in this study is independent of the other HRGC sequences as the sequences were collected from different locations/HRGCs or the same location but at different speeds.

*Data Preprocessing*

Profile data from the GPS sensor was obtained by normalizing the GPS altitude data by its first value. Subsequently, the sequence obtained from the IMU-GPS sensors and the walking profiler were aligned together using the peak of two profiles. Thereafter, an equal distance of IMU-GPS data and walking profiler data was considered on either side of the peak, and the two sequences were merged to form one sequence with 8 parameters, 7 parameters from the IMU-GPS dataset, and one parameter from the walking profiler. **Fig. 4** shows the synchronized profile data obtained from the walking profiler and the GPS sensor. Significant height differences were observed between the two measured profiles at the crossings, especially those near the railway track.

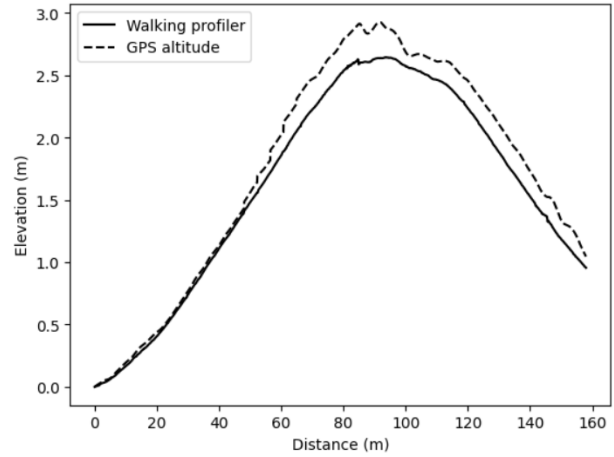

**Fig. 4.** Comparison of profile (Walking Profiler vs GPS)

*Dataset Augmentation*

Deep learning models can approximate very complex functions or natural phenomena, but for developing these models, a dataset containing a very large number of data points is required. Hence, data augmentation was performed using two techniques for creating the training, validation, and test datasets, whose schematic procedure is shown in **Fig. 5**. In the first technique, 5334 sequences were created by inserting random noise into the GPS-measured profile data of the 127 HRGC sequences. The random noise was created using a truncated normal distribution of zero mean, and the standard deviation of the normal distribution was four percent of the difference between the highest and lowest values of the profile data of the sequence. In the second technique, random noise was inserted into the profile data of the HRGC sequence, and subsequently, the sequences were down-sampled into two sequences, one sequence containing odd data points and one containing even data points, using this technique, 5334 sequences were created from 127 HRGC sequences. The number of sequences used for training, validation, and testing were 7888, 1393, and 1387, respectively.

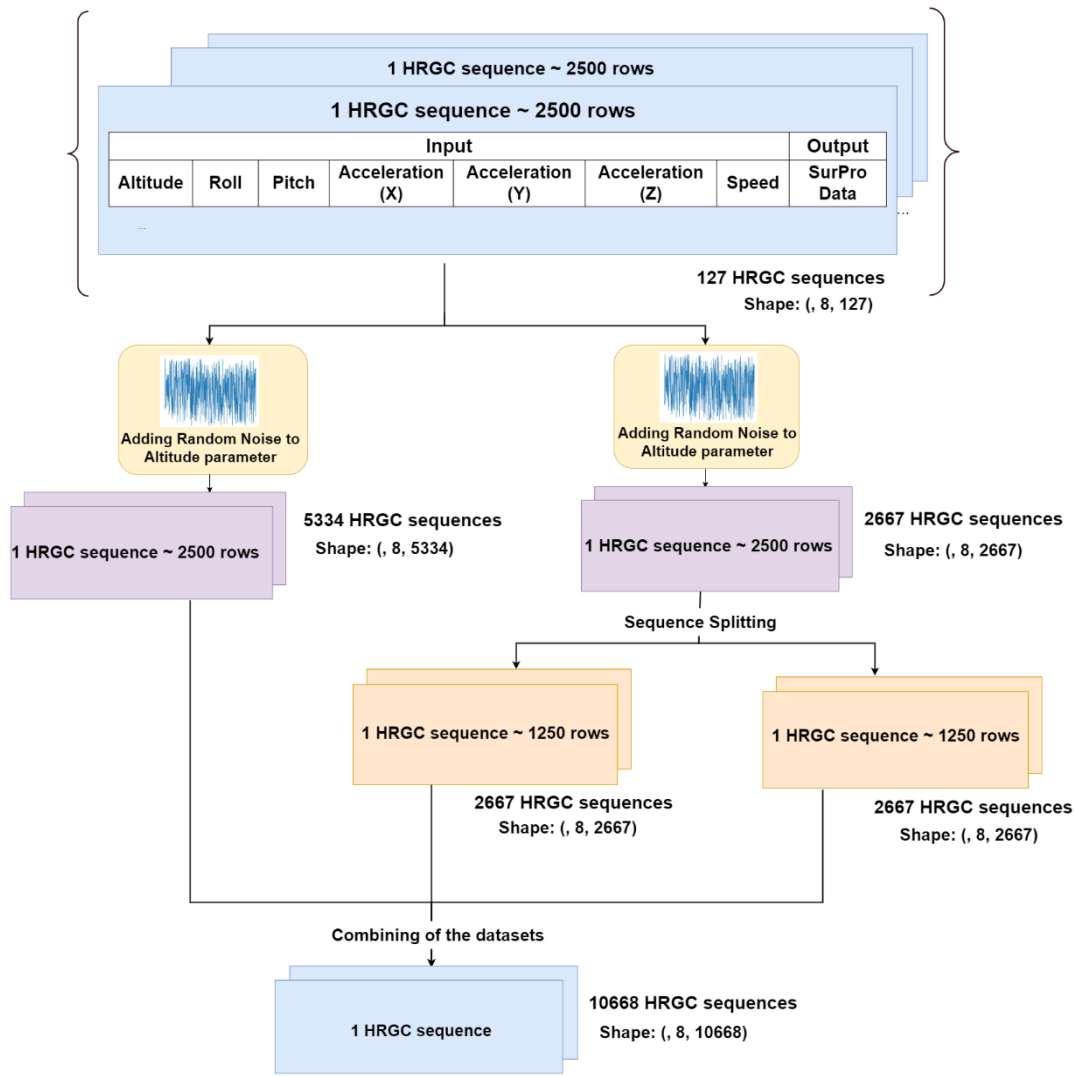

**Fig. 5.** Data augmentation for training, testing, and validation dataset

## METHODOLOGIES

*Sequence-to-Sequence Deep Learning Model*

Sequence-to-sequence deep learning model is a special type of model that deals with data having some sort of dependency on its values at previous time steps. In modern times, sequence-to-sequence deep learning models have gained popularity in several emerging fields such as machine translation and time series data forecasting.

Neural networks that deal with sequence-to-sequence data receive one sequence as the input and deliver another sequence as the output. In earlier times, the size of the neural networks was restricted due to computational and algorithmic challenges. However, recent improvements in more effective algorithms and advanced computational resources such as the graphics processing unit have facilitated applications of neural networks with very large dimensions (De Donno et al. 2010) for sequence-to-sequence modeling.

In this research, sequence-to-sequence hybrid deep learning models were developed using Transformer architecture and LSTM architecture for determining the profile of HRGC using instrumentation data. Both of these deep learning architectures have emerged as vital techniques in dealing with sequential data. Transformers are more effective as compared to the LSTM model (Islam et al. 2023) in dealing with data having long-term dependencies, while with regard to handling data having dominant local features, LSTM architecture performs better than Transformer architecture (Zheng et al. 2020). The strength of both architectures can therefore be leveraged by developing a hybrid model using Transformer and LSTM architecture together. The process of development of hybrid models and related architectures is presented in the subsection hybrid models. Before moving into the development of hybrid models, it is noteworthy to present some information about the LSTM architecture and Transformer architecture.

*Long Short-Term Memory(LSTM) Architecture*

The long Short-Term Memory model represents a kind of recurrent neural network designed by Hochreiter and Schmidhuber. (1997) and is very well recognized for sequence-to-sequence modeling. The main component of the LSTM architecture is the memory cell which can save information over some period of time. The memory cell is composed of three gates namely: (a) forget gate, (b) input gate, and (c) output gate, and these gates regulate the extent of information to be excluded, added, and generated as output from the memory cell. **Fig. 6** shows the different gates of the memory cell and the framework of the LSTM network.

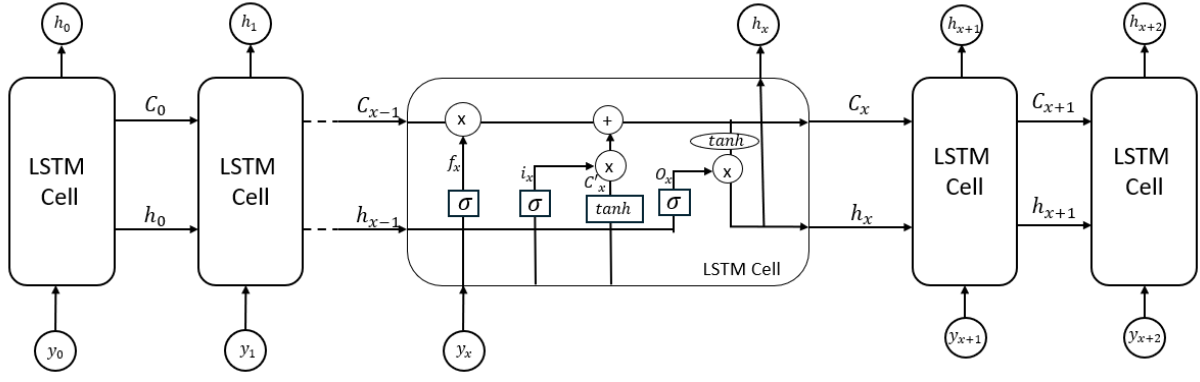

**Fig. 6.** Structure of LSTM network for a distance 0 to x+2

The forget gate in the memory cell of the LSTM executes the function of discarding nonessential information from the state cell. The forget gate takes two inputs: (a) the hidden state of the preceding cell ($h_{x-1}$) and (b) input at a specific position ($y_x$). Eq. (1) shows mathematical computation performed in this gate and Eq. (2) represents the sigmoid function.

$$f_x = \sigma(W_f \cdot [h_{x-1}, y_x] + b_f) \tag{1}$$

$$\sigma(t) = \frac{1}{1+e^{-t}} \tag{2}$$

Here, $W_f$ represents the weight matrix of the forget cell, $[h_{x-1}, y_x]$ represents the concatenation of the input at the current position and the hidden state of the preceding cell, $b_f$ represents the bias vector, $\sigma$ represents the sigmoid activation function, $f_x$ represents the forget gate parameter, and $t$ represents the variable under consideration.

The input gate in the memory cell performs the function of controlling the amount of new information to be included in the cell. The input gate parameter, $i_x$ is presented in Eq. (3). After the computation of $i_x$, computation of candidate cell state is performed, and Eq. (4) represents the mathematical computation of $C'_x$. Thereafter, the computation of cell state is performed, and Eq. (6) represents the mathematical computation.

$$i_x = \sigma(W_i \cdot [h_{x-1}, y_x] + b_i) \tag{3}$$

$$C'_x = tanh(W_c \cdot [h_{x-1}, y_x] + b_c) \tag{4}$$

$$tanh(t) = \frac{e^t - e^{-t}}{e^t + e^{-t}} \tag{5}$$

$$C_x = f_x \times C_{x-1} + i_x \times C'_x \tag{6}$$

Here, $W_i$ and $W_c$ represent the weight matrices of the input gate and candidate cell state, respectively, $b_i$ and $b_c$ represent the bias vectors of the input gate and candidate cell state, respectively, $tanh$ represents the activation function, and $C_x$ represents the updated cell state.

The output gate in the memory cell accomplishes the task of controlling the extent of information that will be delivered to the hidden state of the following cell. Eq. (7) shows the computation of the output gate parameter $O_x$ and Eq. (8) shows the computation of the hidden state of the cell.

$$O_x = \sigma(W_o.[h_{x-1}, y_x] + b_o) \tag{7}$$

$$h_x = O_x \times \tanh(C_x) \tag{8}$$

Here, $W_o$ represents the weight matrix of the output gate and $b_o$ represents the bias vector of the output gate.

*Transformer Architecture*

Transformer is a type of deep learning architecture developed by Vaswani et al. (2017) for machine translation tasks. It is a popular deep learning architecture that can solve the limitations of conventional sequence-to-sequence models such as the LSTM because it can capture long-term dependencies in the sequence. Also, the Transformer models are faster than LSTM because Transformers can do parallel processing of input data unlike the LSTM, which performs sequential processing of input data (Vaswani et al. 2017, Zheng et al. 2020, Islam et al. 2023).

The Transformer architecture mentioned in the paper by Vaswani et al. (2017) has two main blocks: Encoder and Decoder. However, the Transformer architecture in this research was implemented by employing the encoder block only. **Fig. 7** shows a schematic representation of the Transformer architecture (encoder block and positional encoding). Anik et al. (2024) also adopted an encoder-only Transformer architecture in their research. The following two sections discuss positional encoding and different components of the encoder block.

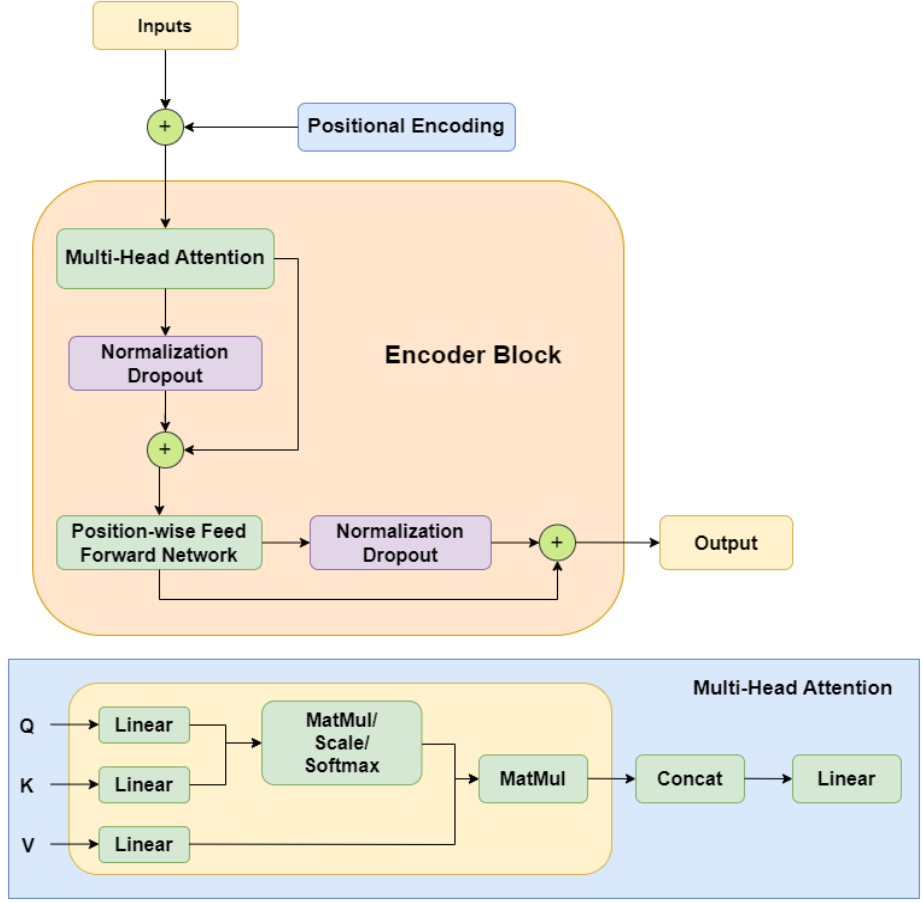

**Fig. 7**. Schematic representation of the Transformer architecture implemented in this research

The Transformer model does not have positional information on different data points in the input sequence due to a lack of convolutional or recurrent operations. However, in sequence-to-sequence modeling, positional information on the data points is essential. Therefore, positional encoding provided the Transformer model with positional information of different dataset points (Vaswani et al. 2017). The mathematical definition of the positional encoding used in this research is presented in Eq. (9) and Eq. (10).

$$PE(position, 2j) = sin\,(position/10000^{2j/d}) \qquad (9)$$

$$PE(position, 2j + 1) = cos\,(position/10000^{2j/d}) \qquad (10)$$

Here, $position$ is the token in the input sequence, $j$ is the dimension of the position index vector and $d$ is the dimension of the representation.

Two main components of the encoder blocks are the Multi-head Self-Attention layer and the position-wise feed-forward network, and some of the other components are the dropout layer, residual connections, and layer normalizations. The multi-head self-attention facilitates capturing the relationship between different parts of the sequence, which was computed by concatenating the output vectors from different single-head self-attention and subsequently performing linear transformation of the concatenated value. The single self-attention head is mathematically defined as follows:

$$Attention(Z) = \sigma(\frac{QK^T}{\sqrt{d_k}})V \tag{11}$$

$$Q = ZW_q; K = ZW_k; V = ZW_v \tag{12}$$

Here $Z$ is the position encoded input, $\sigma$ represents the Softmax function; $Q, K,$ and $V$ represents the query, key, and value, respectively, $d_k$ represents the dimension of the key, and $W_q$, $W_k$, and $W_v$ represents the weight matrices, and T represents the transpose matrices.

The other important component of the encoder block is the position-wise feed-forward layer. This layer transforms the output obtained from the multi-head self-attention layer, facilitating the model to detect more intricate patterns in the dataset. The same network is applied individually to every position of the input sequence. Mathematically, it is defined as follows:

$$FeedForwardNetwork(z_i) = W_2.ReLU(z_i.W_1 + b_1) + b_2 \tag{13}$$

Here $z_i$ is the input vector at position $i$, $W_1$ and $W_2$ represents the weight matrix of the first and second linear layer respectively, $b_1$ and $b_2$ represents the bias vector of the first and second linear layers respectively, and $ReLU$ is the rectified linear unit activation function.

The dropout layers facilitate the model development process by stopping the model from overfitting. The residual connections assist the model development process by addressing the issue of vanishing gradient. Layer normalization helps in stabilizing the training of the network.

# HYBRID LSTM-TRANSFORMER MODELS

## Hybrid Model Configuration

Three different model configurations were considered to develop an optimum sequence-to-sequence hybrid deep learning model leveraging the strengths of LSTM and Transformer architecture. In the first configuration (Transformer LSTM sequential model or model 1), the Transformer layers were applied prior to the LSTM layer, so that the Transformer would capture the global feature in the data using self-attention and subsequently, the LSTM would capture the temporal features in the data using sequential processing. For the second configuration (LSTM Transformer sequential model or model 2), the LSTM layer was applied sequentially before the Transformer layers, and the idea was that the LSTM would first capture the temporal features of the data, followed by the Transformer layer, which would refine this using the self-attention mechanism. Finally, in the third model configuration (LSTM Transformer parallel model or model 3), the LSTM and Transformer layers were placed in parallel. In this model, the LSTM layer and the Transformer layer would independently work on the data, LSTM would capture the local feature in the data through its sequential processing, and the Transformer would capture the global features of the data using a self-attention mechanism. Concatenating the output from the LSTM and Transformer layers would help the model capture the global as well as local context of the data. **Fig. 8** shows the architectures of the three proposed models.

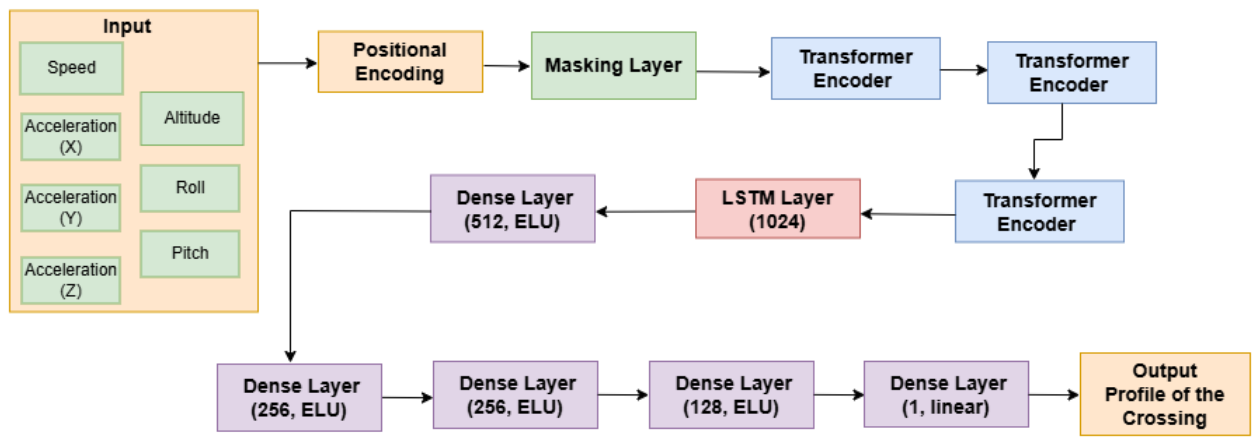

(a)

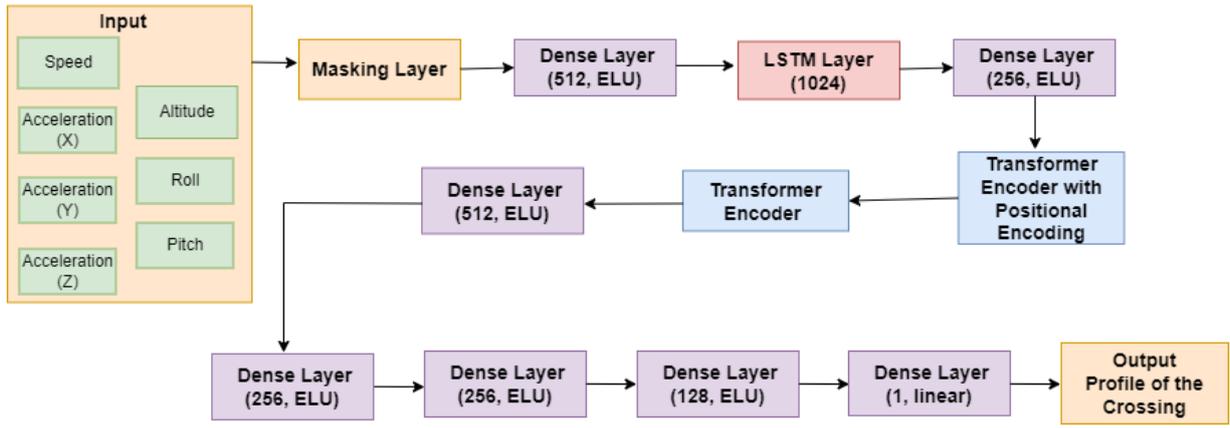

(b)

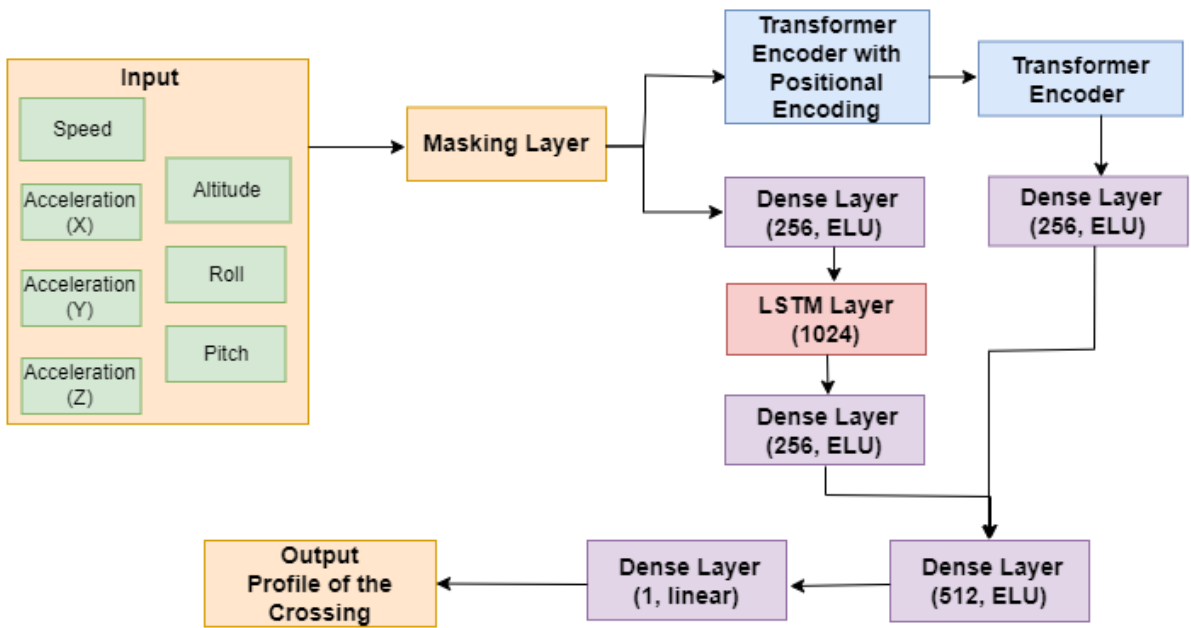

(c)

**Fig. 8.** Model architectures (a) Transformer LSTM sequential model (Model 1), (b) LSTM Transformer sequential model (Model 2), (c) LSTM Transformer parallel model (Model 3)

*Model Development*

The hybrid models were developed using the TensorFlow (Abadi et al. 2016) package in Python. During model development, the different hyperparameters, including learning rate, batch size, and number of parameters of the LSTM layer, were tuned. Also, to prevent overfitting of models, regularization techniques in the form of dropout were employed. However, this caused training instability, so dropout was removed from the model development technique. The hyperparameters of the final developed models are shown in **Table 1**.

Table 1. Hyperparameters of the final developed models

| Hyperparameters | Model 1 | Model 2 | Model 3 |
|---|---|---|---|
| Learning rate | 0.0001 | 0.0001 | 0.0001 |
| Batch size | 32 | 32 | 32 |
| Number of parameters of the LSTM layer | 1024 | 1024 | 1024 |
| Optimization function | Adam | Adam | Adam |
| Number of heads (transformer encoder) | 2 | 2 | 2 |
| Dimension of feed forward network (transformer encoder) | 1024 | 2048 | 2048 |

*Model Performance Evaluation Matrices*

Root Mean Square Error (RMSE) and Mean Absolute Error (MAE) were selected for evaluating the performance of the model. These metrics are very commonly used to evaluate the performance of regression models. Equations 14 and 15 show the mathematical representation of RMSE and MAE. **Table 2** shows the performance of the model based on these metrics. The performance of model 1 was lower than expected as the RMSE and MAE values were higher, while the performance of model 2 and model 3 met the expectations. After evaluating the performance of the models on the test dataset, the generalizability of the models was also evaluated to test the robustness of the model to data variability and the performance of the model under different conditions.

$$RMSE = \sqrt{\frac{\sum_{i=1}^{N}(y_i - \hat{y}_i)^2}{N}} \quad (14)$$

$$MAE = \frac{\sum_{i=1}^{N}|y_i - \hat{y}_i|}{N} \quad (15)$$

Where $RMSE$ is the root mean square error, $MAE$ is the mean absolute error, $y_i$ is the predicted value from the model, $\hat{y}_i$ is the ground truth data from the walking profiler, and $N$ is the number of data points in the sequence.

**Table 2 Performance of model on training, validation and test dataset**

| Name | Training | | Validation | | Test | |
|---|---|---|---|---|---|---|
| | RMSE (m) | MAE (m) | RMSE (m) | MAE (m) | RMSE (m) | MAE (m) |
| Model 1 | 0.14 | 0.12 | 0.14 | 0.12 | 0.14 | 0.12 |
| Model 2 | 0.05 | 0.04 | 0.05 | 0.04 | 0.05 | 0.04 |
| Model 3 | 0.07 | 0.05 | 0.07 | 0.05 | 0.07 | 0.05 |

*Model Generalizability and Discussions*

The generalizability of the models was evaluated using nine HRGC sequences excluded from the training, validation, and testing datasets. **Table 3** presents the performance of the models. Model 1 demonstrated lower-than-expected performance, whereas models 2 and 3 met expectations. The RMSE and MAE values for models 2 and 3 were comparable to those observed during training, validation, and testing, which indicated their strong generalizability.

To assess the models' performance under scenarios with low-resolution data, the nine HRGC sequences were down-sampled by a factor of 2 to simulate data collection using low-cost instruments with reduced acquisition frequency. As shown in **Table 3**, models 2 and 3 performed satisfactorily on the down-sampled dataset and maintained their ability to accurately determine HRGC profiles. Conversely, model 1's performance remained below expectations and thus rendering it unsuitable for this application aiming to identify hump crossings.

**Table 3. Results: Generalization of model**

| Down sampling factor | Models | RMSE (m) | MAE (m) |
|---|---|---|---|
| - | Model 1 | 0.13 | 0.11 |
| | Model 2 | 0.09 | 0.07 |
| | Model 3 | 0.07 | 0.06 |
| 2 | Model 1 | 0.15 | 0.13 |
| | Model 2 | 0.08 | 0.07 |
| | Model 3 | 0.07 | 0.06 |

Among the models, both 2 and 3 demonstrated the ability to reliably determine HRGC profiles. However, model 3 holds a slight advantage over model 2 due to its lower number of trainable parameters (5,965,169 for model 3 compared to 13,235,713 for model 2), making it more computationally efficient. Finally, models 2 and 3 were deployed to generate estimated HRGC profiles. **Fig. 9** illustrates the profiles determined by the deep learning models, which closely matched the ground truth data and confirmed their efficiency and accuracy in profile determination.

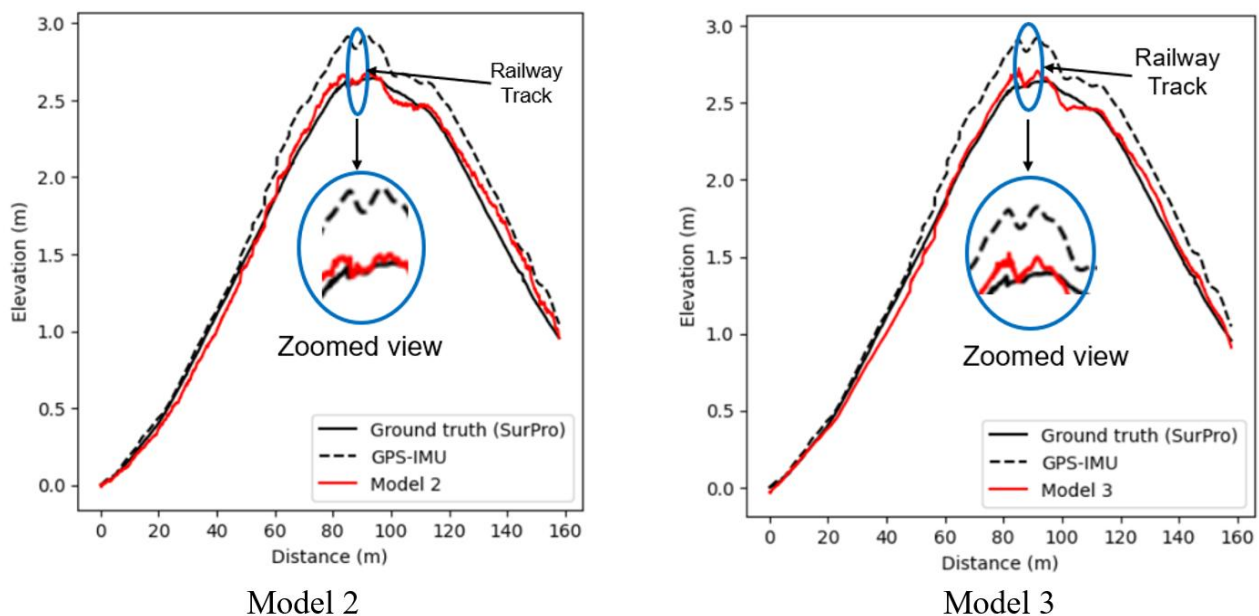

Model 2            Model 3

**Fig. 9.** Profile from deep learning models

**CONCLUSIONS**

This study demonstrated the effectiveness of hybrid deep learning models for determining the profiles of Highway Railway Grade Crossings (HRGCs) using instrumentation data collected from IMU and GPS sensors. Field data were obtained from various HRGCs along the Red Rock Corridor in Oklahoma, USA. Ground truth data were collected using a walking profiler at a speed of 3.0 km/hr, while IMU and GPS data were captured at speeds ranging from 10 km/hr to 32 km/hr.

    Three hybrid deep learning model configurations were developed to estimate HRGC profiles using instrumentation data. During the model development and deployment phases, the input parameters included vehicle acceleration along three axes, roll, pitch, vehicle speed, and GPS-measured HRGC profiles. Model 1 exhibited lower-than-expected performance, while models 2 and 3 met expectations on training, validation, and testing datasets. Further evaluation of generalizability using nine additional HRGC sequences showed a similar performance trend. When tested on low-resolution data, models 2 and 3 maintained comparable performance to the original datasets, which

further confirmed their robustness. Ultimately, models 2 and 3 were deployed to generate HRGC profiles, which closely aligned with the ground truth profiles. This validates the sequence-to-sequence hybrid deep learning models as reliable tools for determining HRGC profiles, such that hump crossings can be identified for enhanced HRGC safety.

## DATA AVAILABILITY

All data, tools, or models that support the findings of this study are available from the corresponding author upon reasonable request.

## ACKNOWLEDGMENT

The data used herein were collected under the SP&R 2296 research project, "Highway/Rail Intersection Hump or High-Profile Crossings Problems", sponsored by the Oklahoma Department of Transportation (ODOT).

## REFERENCES

ABC News. 2021. "Dramatic video shows Amtrak train slamming into semi-truck car hauler" Accessed December 17, 2024. https://abcnews.go.com/US/dramatic-video-shows-amtrak-train-slamming-semi-truck/story?id=80619175

Anik, B. T. H., Z. Islam, and M. Abdel-Aty. 2024. "A time-embedded attention-based Transformer for crash likelihood prediction at intersections using connected vehicle data." *Transportation Research Part C: Emerging Technologies*, *169*, 104831. https://doi.org/10.1016/j.trc.2024.104831

AASTHO. 2018. "A Policy on Geometric Design of Highways and Streets", 7th Edition. American Association of State Highway Transportation Officials, Washington D.C.

Abadi, M., Barham, P., Chen, J., Chen, Z., Davis, A., Dean, J., ... & Zheng, X. 2016. "{TensorFlow}: a system for {Large-Scale} machine learning. In *12th USENIX symposium on operating systems design and implementation (OSDI 16)*, pp. 265-283.

AREMA (American Railway Engineering Maintenance-of-Way Association) (2018), "*2018 Manual*

for Railway Engineering", AREMA, Lanham, MD.

Bae, H. S., J. Ryu, and J. C. Gerdes. (2001, August). "Road grade and vehicle parameter estimation for longitudinal control using GPS". In *Proceedings of the IEEE Conference on Intelligent Transportation Systems*, pp. 25-29.

Bosman, T. (2015). "Evaluating Grade and Cross Slope of a Road Using Inertial Sensors and GPS."

Chatterjee, K., D. Vivanco, X. Yang, and J. Q. Li. 2024. "Enhancing Pavement Performance through Balanced Mix Design: A Comprehensive Field Study in Oklahoma." In *International Conference on Transportation and Development 2024*. Atlanta GA.
https://doi.org/10.1061/9780784485538.045

De Donno, D., A. Esposito, L. Tarricone, and L. Catarinucci. 2010. "Introduction to GPU computing and CUDA programming: A case study on FDTD [EM programmer's notebook]." *IEEE Antennas and Propagation Magazine*, *52*(3), pp: 116-122.
https://doi.org/10.1109/MAP.2010.5586593

Eck, R. W., & Kang, S. K. (1991). "Low-Clearance Vehicles at Rail-Highway Grade Crossings: An Overview of the Problem and Potential Solutions." *Transportation Research Record*, Vol. 1327, pp: 27-35.

Eck, R. W., & Kang, S. K. (1992). "Roadway design standards to accommodate low-clearance vehicles (with discussion and closure)." *Transportation Research Record*, Vol. 1356, pp: 80-89.

Guang, X., Y. Gao, P. Liu, and G. Li. 2021. "IMU data and GPS position information direct fusion based on LSTM". Sensors, 21(7), 2500. https://doi.org/10.3390/s21072500

Hochreiter, S., and J. Schmidhuber. 1997. "Long short-term memory". Neural Computation, 9(8), pp: 1735-1780. https://doi.org/10.1162/neco.1997.9.8.1735

Islam, S., H. Elmekki, A. Elsebai, J. Bentahar, N. Drawel, G. Rjoub, and W. Pedrycz. 2023. "A comprehensive survey on applications of Transformers for deep learning tasks." Expert Systems with Applications, 122666. https://doi.org/10.1016/j.eswa.2023.122666

Jauch, J., J. Masino, T. Staiger, and F. Gauterin. 2017. "Road grade estimation with vehicle-based inertial measurement unit and orientation filter." IEEE Sensors Journal, 18(2), pp: 781-789. https://doi.org/10.1109/JSEN.2017.2772305

Khattak, A. J., Tang, Z., & Lee, M. (2015). "*Application of Light Detection and Ranging Technology to Assess Safe Passage of Low Ground Clearance Vehicles at Highway-Rail Grade Crossings*" (No. 26-1121-0018-002). Mid America Transportation Center. Lincoln, NE.

Lee, M., and A. J. Khattak. 2016. "Application of Light Detection and Ranging Technology to Assess Safe Passage of Low-Ground-Clearance Vehicles at Highway-Rail Grade Crossings." Transportation Research Record, 2545(1), pp: 131-139. https://doi.org/10.3141/2545-14

Liu, Q., T. Wang, and R. R. Souleyrette. 2017. "A 3D evaluation method for rail–highway hump crossings." *Computer-Aided Civil and Infrastructure Engineering*, *32*(2), pp: 124-137. https://doi.org/10.1111/mice.12244

Mtabazi, M., & Russell, E. R. (2003). "*Identification of Hump Highway/rail Crossings in Kansas*" Report No. K-TRAN: KSU-99-1). Kansas Department of Transportation, Topeka KS.

Ogden, B. D., and C. Cooper. 2019. "Highway-rail crossing handbook." Federal Highway Administration. Washington, D.C.

Vaswani, A. (2017). "Attention is all you need". The 31st Conference on Neural Information Processing Systems (NIPS 2017), Long Beach CA.

Zheng, Y., X. Li, F. Xie, and L. Lu. 2020, May. "Improving end-to-end speech synthesis with local recurrent neural network enhanced Transformer." In *ICASSP 2020-2020 IEEE International Conference on Acoustics, Speech and Signal Processing (ICASSP)*. IEEE. pp. 6734-6738. Barcelona, Spain. https://doi.org/10.1109/ICASSP40776.2020.9054148